\documentclass[letterpaper, 10 pt, conference]{ieeeconf}  
\usepackage{amsmath,amsfonts}
\usepackage{fancyhdr}
\usepackage{algorithmic}
\usepackage{array}
\usepackage[caption=false,font=normalsize,labelfont=sf,textfont=sf]{subfig}
\usepackage{textcomp}
\usepackage{stfloats}
\usepackage{url}
\usepackage{enumerate}
\usepackage{verbatim}
\usepackage{graphicx}
\usepackage{cite}
\usepackage{booktabs,makecell, multirow, tabularx, pifont}
\usepackage{booktabs}
\usepackage[linesnumbered,lined,boxed,commentsnumbered,ruled]{algorithm2e}
\usepackage{marginnote}
\usepackage{soul}
\usepackage{bigstrut}
\usepackage{diagbox}
\usepackage{stackengine}
\usepackage{siunitx}

\IEEEoverridecommandlockouts                              

\title{\LARGE \bf
A Decapod Robot with Rotary Bellows-Enclosed Soft Transmissions
}

\author{Yiming~He$^{1}$,
        Yuchen~Wang$^{2}$,
        Yunjia~Zhang$^{1}$,
        and Shuguang~Li$^{1,3*}$
\thanks{$^{1}$Yiming He, Yunjia Zhang, and Shuguang Li are with the Department of Mechanical Engineering, Tsinghua University, Beijing, 100084, China.}%
\thanks{$^{2}$Yuchen Wang is with the Weiyang College, Tsinghua University, Beijing, 100084, China.}%
\thanks{$^{3}$Shuguang Li is with Beijing Key Laboratory of Transformative High-end Manufacturing Equipment and Technology, Tsinghua University, Beijing 100084, China.}%
\thanks{\textsuperscript{*}Corresponding author: Shuguang Li (email:lisglab@tsinghua.edu.cn).}%
}

\begin{document}

\maketitle
\thispagestyle{fancy}
\pagestyle{empty}

\lhead{}
\lfoot{}
\cfoot{\small{Copyright \copyright 2025 IEEE. Personal use of this material is permitted. \\
		However, permission to use this material for any other purposes must be obtained 
		from the IEEE\\by sending an email to pubs-permissions@ieee.org.}}
\rfoot{}
\renewcommand{\headrulewidth}{0mm}


\begin{abstract}

Soft crawling robots exhibit efficient locomotion across various terrains and demonstrate robustness to diverse environmental conditions. Here, we propose a valveless soft-legged robot that integrates a pair of rotary bellows-enclosed soft transmission systems (R-BESTS).
The proposed R-BESTS can directly transmit the servo rotation into leg swing motion. A timing belt controls the pair of R-BESTS to maintain synchronous rotation in opposite phases, realizing alternating tripod gaits of walking and turning.
We explored several designs to understand the role of a reinforcement skeleton in twisting the R-BESTS' input bellows units.
The bending sequences of the robot legs are controlled through structural design for the output bellows units.
Finally, we demonstrate untethered locomotion with the soft robotic decapod.
Experimental results show that our robot can walk at 1.75 centimeters per second (0.07 body length per second) for 90 min, turn with a 15-centimeter (0.6 BL) radius, carry a payload of 200 g, and adapt to different terrains.

\end{abstract}

\section{INTRODUCTION}

Soft robots can adapt to complex environments and safely interact with people and surroundings\cite{r1}. These remarkable properties have led to widespread adoption in grasping\cite{r6}, rehabilitation devices\cite{r7}, wearables\cite{r8}, and mobile aqueous\cite{r4}, airborne\cite{r9}, and ground robots\cite{r10}. Over the last decade, several soft robots inspired by legged animals in nature have emerged, ranging from quadruped\cite{r2,r5,r15,r16,r25}, hexapod\cite{r11}, spider\cite{r13} to multi-legged\cite{r3,r14} robots.

As for soft crawling robots, He et al. \cite{r18} created a soft tethered autonomous robot without electronics based on intelligent mechanical metamaterials and embodied logic.
The robot body consisted of a pneumatic actuator and a kirigami-inspired architecture, which served as a flexible and convenient platform for the reconfigurable, modular control units. The control modules constrained the local rotation of the kirigami when they were activated by stimuli-responsive materials, such as liquid crystal elastomers that respond to heat or light.
Lee et al. \cite{r19} developed a buckling-sheet ring oscillator capable of performing translational and rotational motions over varied terrain.
It could directly generate movement from its interaction with its surroundings.
It could climb upward against gravity and downward against the buoyant force encountered underwater.
Bell et al. \cite{r20} developed a robot inspired by the tube feet in echinoderms. 
The robot rotated by actuating or retracting one of the four sets of tube feet. It was capable of rolling on ferrous surfaces with a pull-and-roll technique.
Drotman et al. \cite{r21,r22} presented an autonomous, electronics-free walking robot.
The robot only required a constant pressurized air source to power the control and actuation systems. The pneumatic control circuit could generate walking gaits for a soft-legged quadruped with three degrees of freedom per leg. It could also switch between gaits to control the direction of locomotion.
MacCurdy et al. \cite{r27} described the design and use of bellows actuators as basic force transfer elements for printing hydraulic robots, and showcased their use in a hexapod robot.

Pneumatically-driven soft crawling robots generally rely on external pneumatic circuits, such as intricate valving arrays or multiple external pressure sources, increasing the overall complexity, weight, and cost of the system. These multi-valve arrangements also complicate the control strategy, as each valve must be independently actuated and coordinated. 
Designing soft robots based on fluidic logic can elegantly reduce robot control complexity.

\begin{figure}[t]
  \centering
  \includegraphics[totalheight=2.0 in]{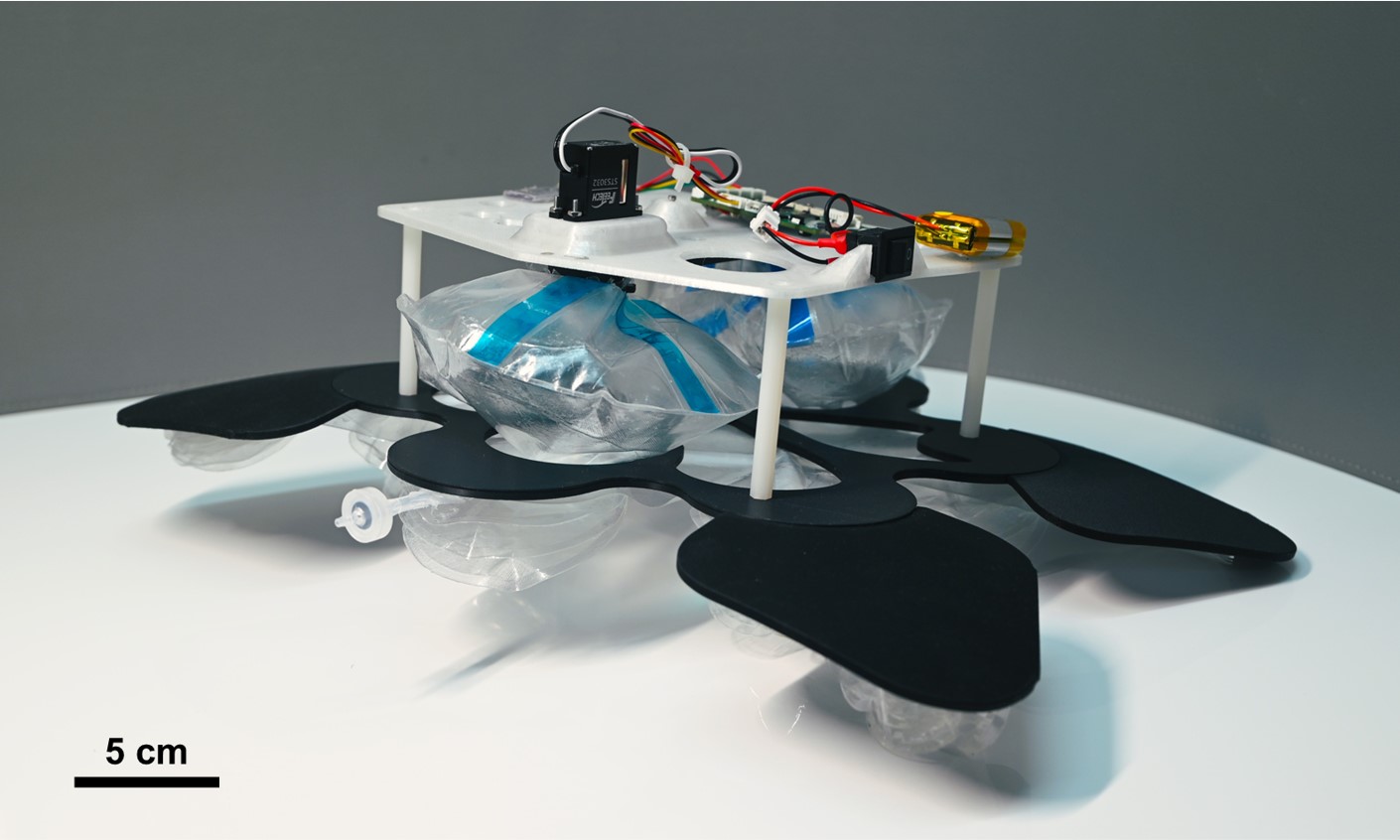}
  \caption{The untethered soft decapod robot with the R-BESTS.} \label{Fig:1}
\end{figure}

Wehner et al. \cite{r28} reported the untethered operation of a robot composed solely of soft materials. The robot was controlled with microfluidic logic that autonomously regulated fluid flow and catalytic decomposition of an on-board monopropellant fuel supply.
Jones et al. \cite{r23} employed a simple and versatile fabrication methodology to assemble monolithic actuators programmed using the rules and tools of fluid mechanics.
They demonstrated how to use bubble casting to achieve versatile and programmable actuators.
Vasios et al. \cite{r17} demonstrated a pneumatically-driven four-legged soft robot that could walk using a single input. 
They embedded actuation sequencing in the system by connecting fluidic actuators with narrow tubes to exploit the effects of flow viscosity. 
Zhai et al. \cite{r24} presented an approach for the 3D printing of soft, airtight pneumatic robotic devices with embedded fluidic control components simultaneously.
A pneumatic valve that could be printed without manual post-treatment was designed. Besides, the gripper could use feedback from a contact switch to grip an object and hold it until a gravity switch was activated.
Some researchers had designed an untethered single DOF soft crawling robot \cite{r12}. By leveraging viscous flows within the actuator to produce nonuniform pressure between bladders, the motion of the hexapod was generated with a single pressure input rather than relying on complex feedback control or multiple inputs.

In this study, we design a valveless soft-legged decapod robot. This robot has a pair of rotary bellows-enclosed soft transmission systems (R-BESTS), which can directly transmit the rotation from a servo into the swinging motion of its legs. 
We control the sequences of different legs by designing the radii of the bellows units for each leg.
With the use of a servo motor and a timing belt, the robot realizes alternating tripod gaits and achieves walking and turning movements on different surfaces and terrains.
By eliminating the need for valves and tubing, our approach simplifies the overall robot design and control scheme. Additionally, the enclosed nature of the transmission reduces fluid leakage risk, offering improved reliability and consistent performance over a range of operating conditions.

Compared to similar systems reported in recent literature, which usually require extensive multi-channel or external pneumatic components, our valveless configuration uses fewer moving parts, is easier to assemble, and can be utilized in a variety of soft robotic applications. This approach also opens up possibilities for more compact designs, as the fluidic channels and bellows units can be arranged to suit diverse morphologies without the burden of complex tubing networks.

The main contributions of this work can be summarized as follows:

(1) We introduce a compact, valveless transmission strategy for fluid-driven soft robots.

(2) We demonstrate the controllability and reliability of a single-actuator soft crawling robot through the R-BESTS.

\section{DESIGN AND FABRICATION}

\subsection{Working principle of the R-BESTS}
\begin{figure}[htbp]
  \centering
  \includegraphics[totalheight=2.05in]{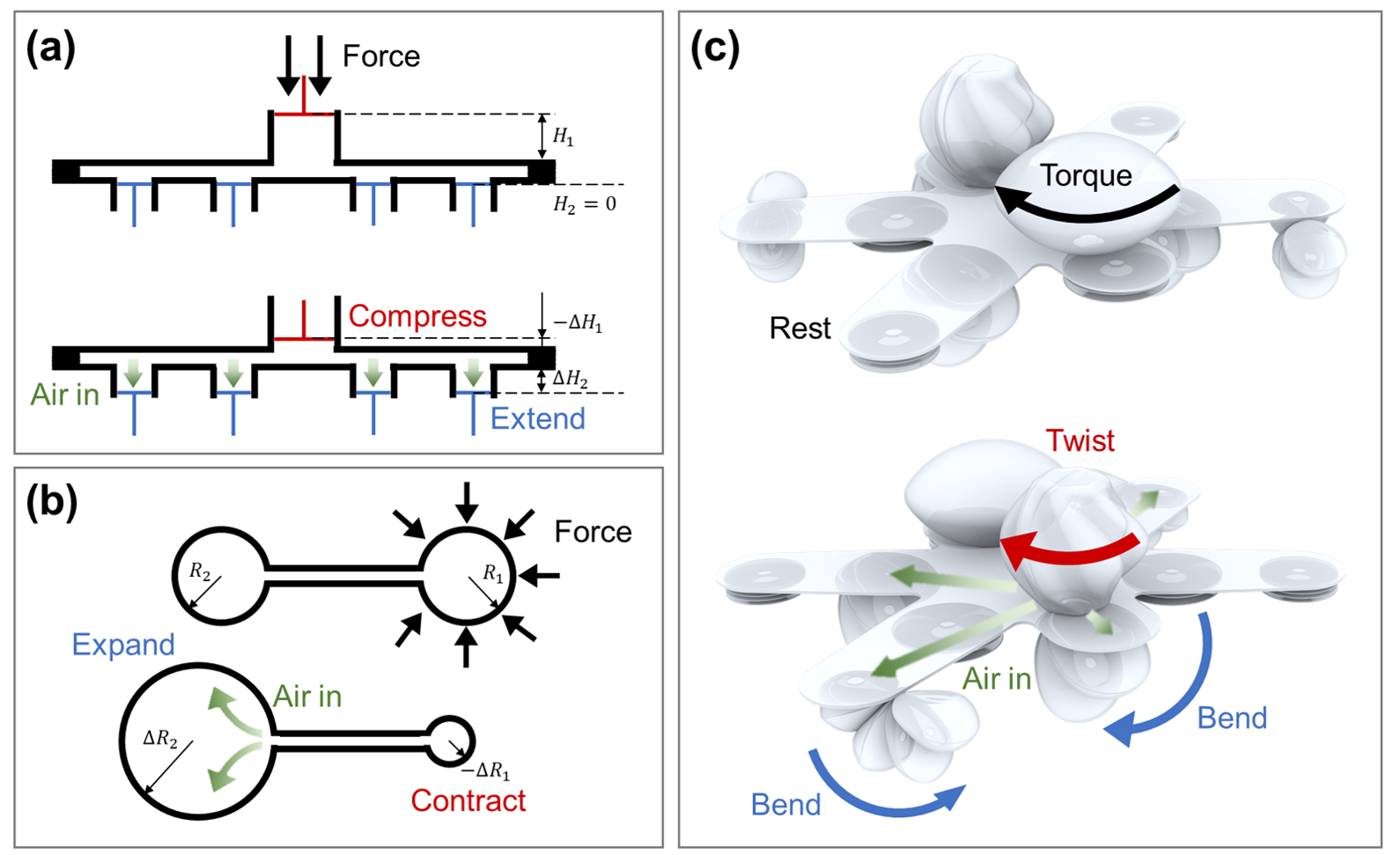}
  \caption{The working principle of a transmission system consisting of piston (a), two connected balloons (b), and the R-BESTS (c).} \label{Fig:2}
\end{figure}

Fluidic actuation is the most commonly used method in soft robotics.
Most existing fluidic actuation methods require pumps to control the actuators, posing a fluid leakage risk. 
However, the closed pneumatic system not only overcomes the leakage problem but also efficiently transmits the motion and pressure. In a communicating vessel (Fig.\ref{Fig:2}(a)), pushing one piston at the input end drives the pistons at the output end to extend. Since the system is closed, pressure can quickly transmit to the other end, reflecting the capability of high bandwidth.
The closed system can also be made of soft materials like rubbers or other films. 
When two balloons are connected, squeezing one will cause the other to expand, as shown in Fig.\ref{Fig:2}(b).

Inspired by the system above, we propose the R-BESTS (Fig.\ref{Fig:2}(c)). When applying torque to a single input bellows unit, it twists, and the air in the system flows from the input to the output, resulting in the bending of output bellows units. This bending motion is achieved by the design of the heat press bonding area. The pair of bellows units have $180^o$ phase differences so that the robot can have an alternating tripod gait.

\subsection{Fabrication of the R-BESTS}

We manufactured the R-BESTS by heat-pressing \SI{0.1}{\milli\meter}-thick films of thermoplastic polyurethane (TPU) and \SI{0.03}{\milli\meter}-thick films of polyethylene terephthalate (PET) coated with hot melt adhesive film. The coated-PET was used as the mask to define where the TPU films were bonded.

\begin{figure}[htbp]
  \centering
  \includegraphics[totalheight=1.7in]{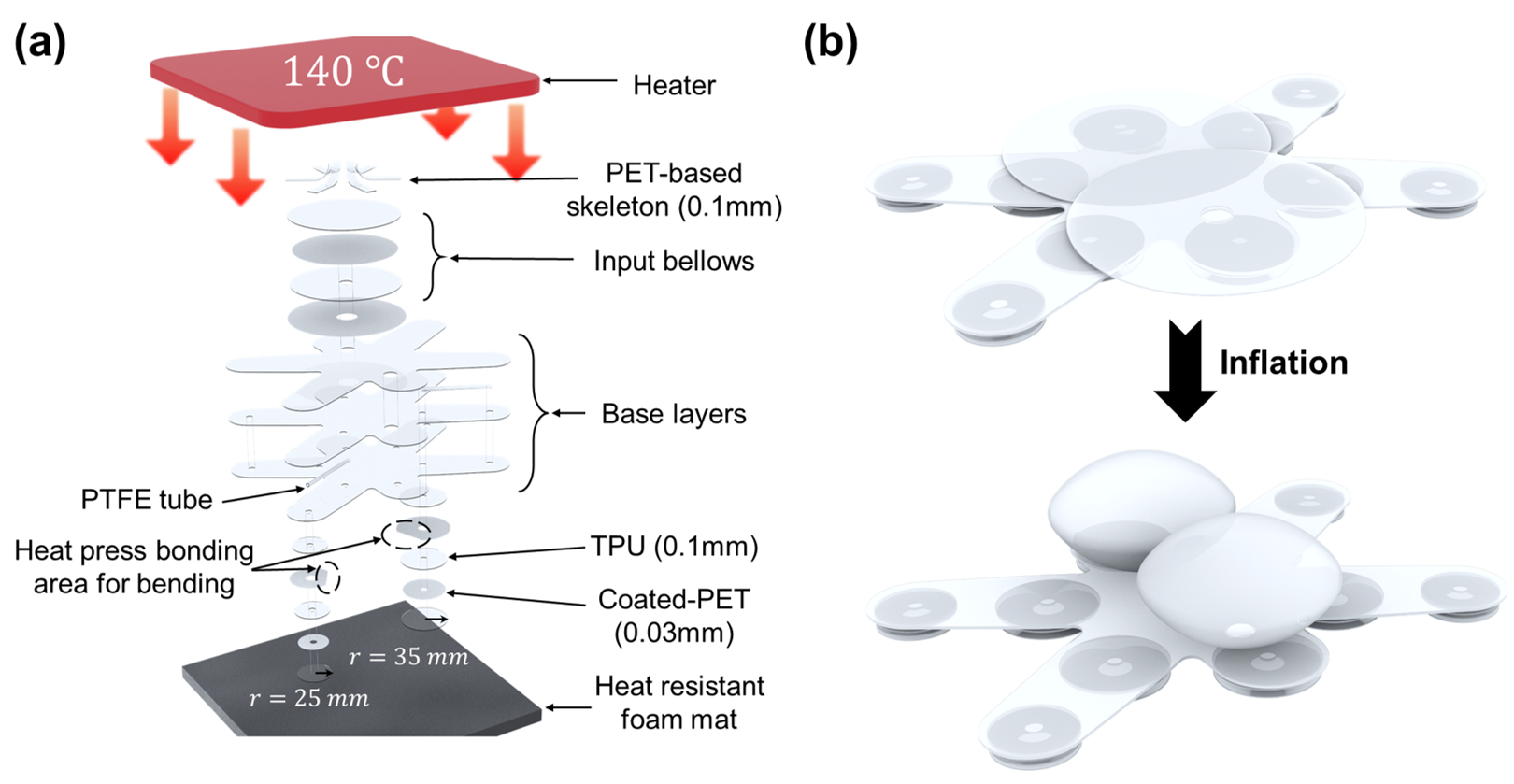}
  \caption{Overview of the fabrication process of the R-BESTS. (a) Three-dimensional schematic diagram. (b) After heat-pressing, the R-BESTS forms a flat laminated structure. Upon inflation, the pair of input bellows units expand.} \label{Fig:3}
\end{figure}

As shown in Fig.\ref{Fig:3}(a), a \SI{0.1}{\milli\meter}-thick PET film was used as a reinforcement skeleton on the top of the input bellows unit. It can produce greater volume change when twisting the input bellows unit than without a skeleton, and it also accelerates the recovery during the rotary process in the reverse direction.
We placed a Polytetrafluoroethylene (PTFE) tube between the two TPU films. The tube not only created a pneumatic channel during heat-pressing but also effectively prevented the film from buckling and jamming the gas channel during the driving process.
At the output end, different bending directions of the R-BESTS are realized by designing the shape and orientation of the coated-PET film.

After placing the laminated structure in a heat-pressing machine at \SI{140}{\celsius} for \SI{150}{\second}, we got a two-dimensional laminated structure, as shown in Fig.\ref{Fig:3}(b). 
Finally, we got the R-BESTS by inflating the pair of closed systems with air, respectively, and sealing the vents. When twisting the input bellows unit, the air is transferred to the output bellows units within the closed system, causing the bending movement of the robot's legs.
Each R-BESTS takes approximately \SI{100}{\minute} to fabricate, including film cutting, assembling, heat-pressing, and post-processing.

\begin{figure}[htbp]
  \centering
  \includegraphics[totalheight=1.8in]{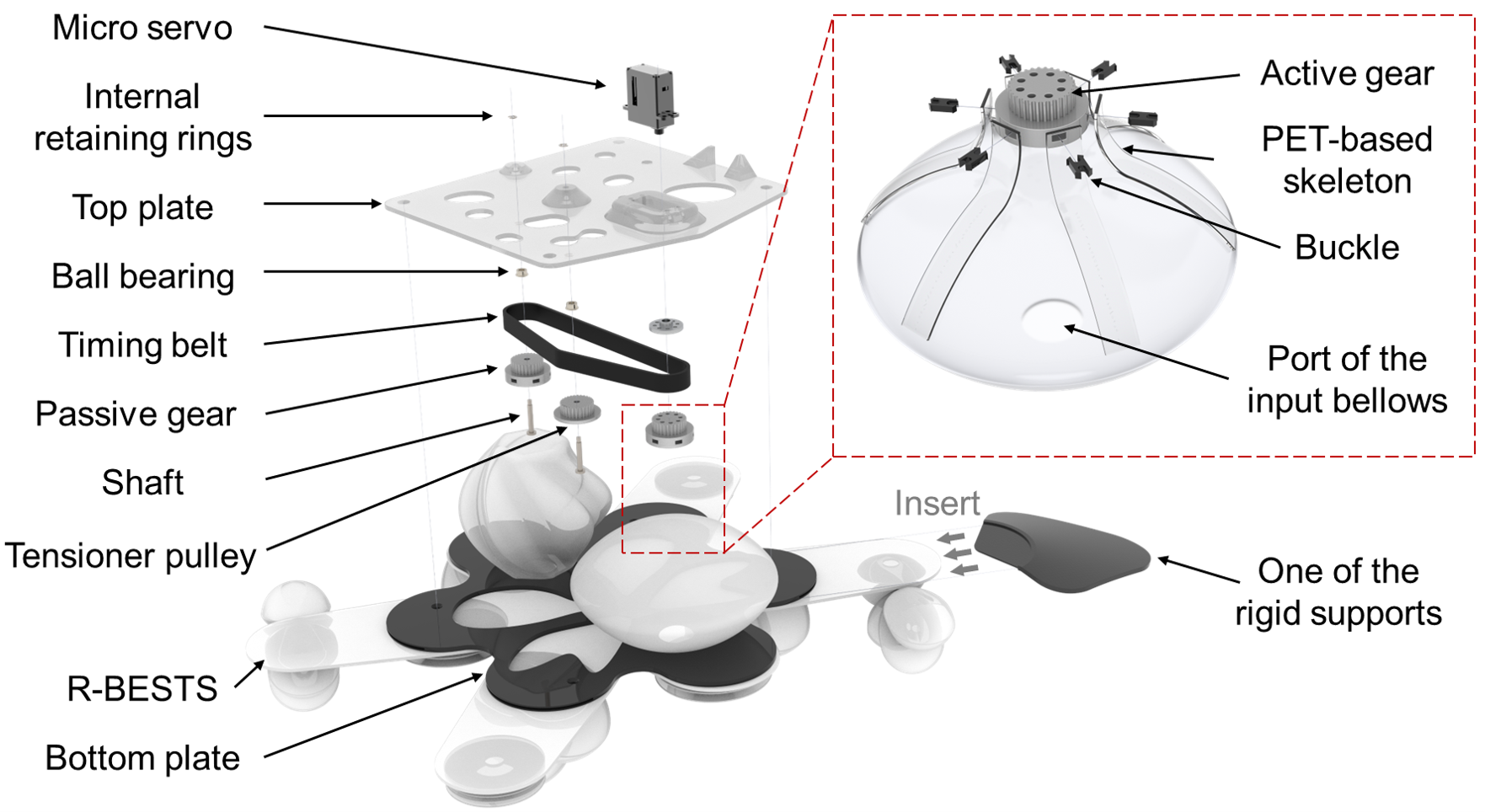}
  \caption{Exploded view of the robot.} \label{Fig:4}
\end{figure}

\subsection{Hardware implementation}

We developed a decapod soft crawling robot, as shown in Fig.\ref{Fig:4}. Only one micro servo serves as input to produce the bending motions of the leg by twisting the R-BESTS. The frictions between the tips of the leg and the ground lead to a walking or turning motion. The top and bottom plates of the robot were 3D printed using Polylactic Acid (PLA) filament. A timing belt was implemented to control the pair of input bellows units through one servo.

During the assembly process, we first turned the phase offset of the pair of input bellows units by rotating the servo 180 degrees. Then, the tensioner pulley was installed to fix the phase difference of the pair of input bellows units. When the servo rotated, the two units of input bellows rotated in the same direction due to the timing belt mechanism. The volume of one input bellows unit decreased while the other increased due to the offset angles. 
This coordinated actuation enables the realization of an alternating tripod gait.

To connect the rigid active gear and the flexible R-BESTS, we made a groove on the side of the gear and inserted the PET-based skeleton on the top of the R-BESTS into the groove through a nylon buckle. When the gears rotated, the input bellows unit was driven to twist through the PET-based skeleton.

\begin{figure}[t]
  \centering
  \includegraphics[totalheight=1.45in]{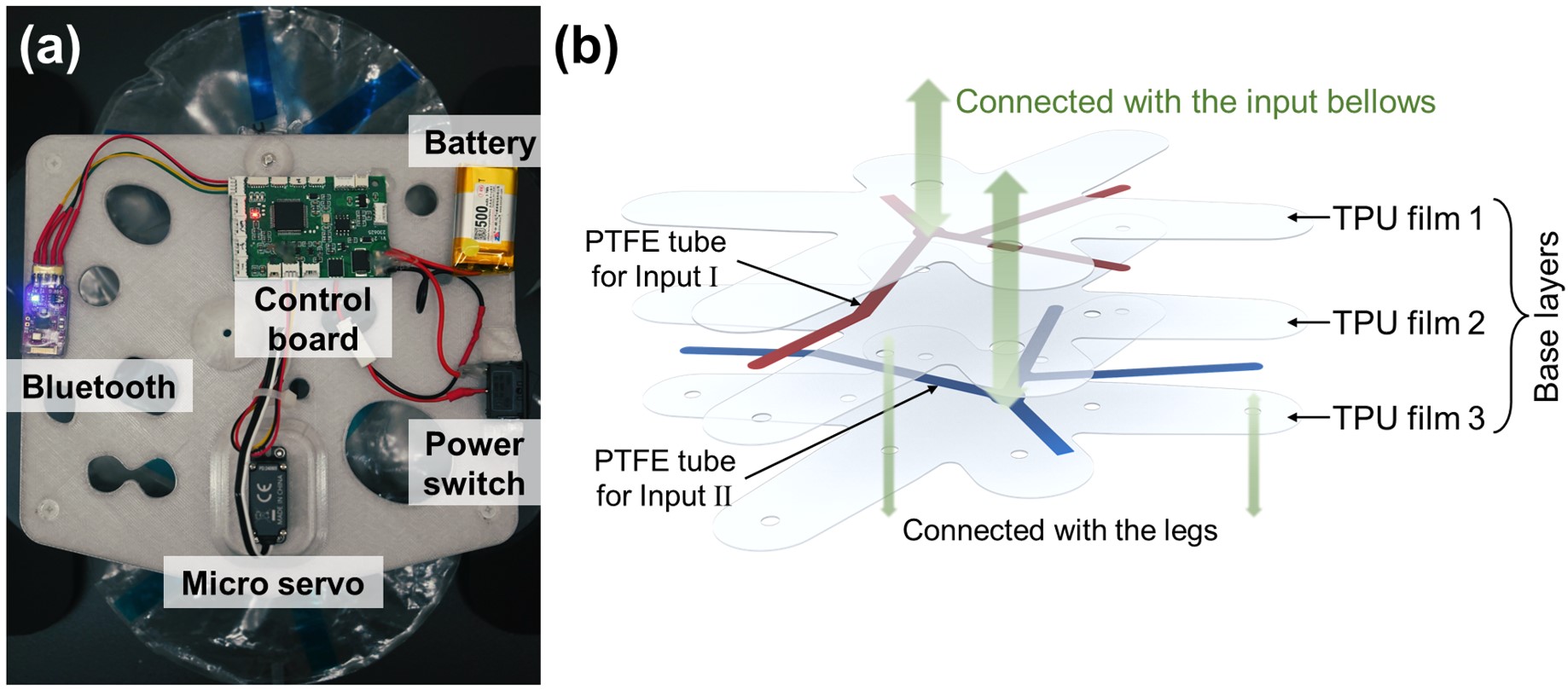}
  \caption{The electronics and pneumatics of the soft crawling robot. (a) Embedded electronic components in the robot. (b) Pneumatic channels for alternating tripod gait.} \label{Fig:5}
\end{figure}

Fig.\ref{Fig:5}(a) shows the embedded electronic components: a 7.4V Lithium-polymer battery (\SI{500}{\milli\ampere\hour}), a Bluetooth module for wireless communication and remote control, an STM32F407 microcontroller as the processing and computing board, and a servo. All electronic components were mounted on the top plate of the robot. 

To integrate the two R-BESTS, we used three layers of TPU film to form the base layers of the R-BESTS, as shown in Fig.\ref{Fig:5}(b). Two sets of PTFE tubes were arranged between two TPU films. In this way, one input bellows unit can control the movement of five units of output bending bellows.

\section{CHARACTERIZATIONS OF THE ROBOT}
\subsection{Control of the actuation sequencing}
\begin{figure}[bp]
  \centering
  \includegraphics[totalheight=2.5in]{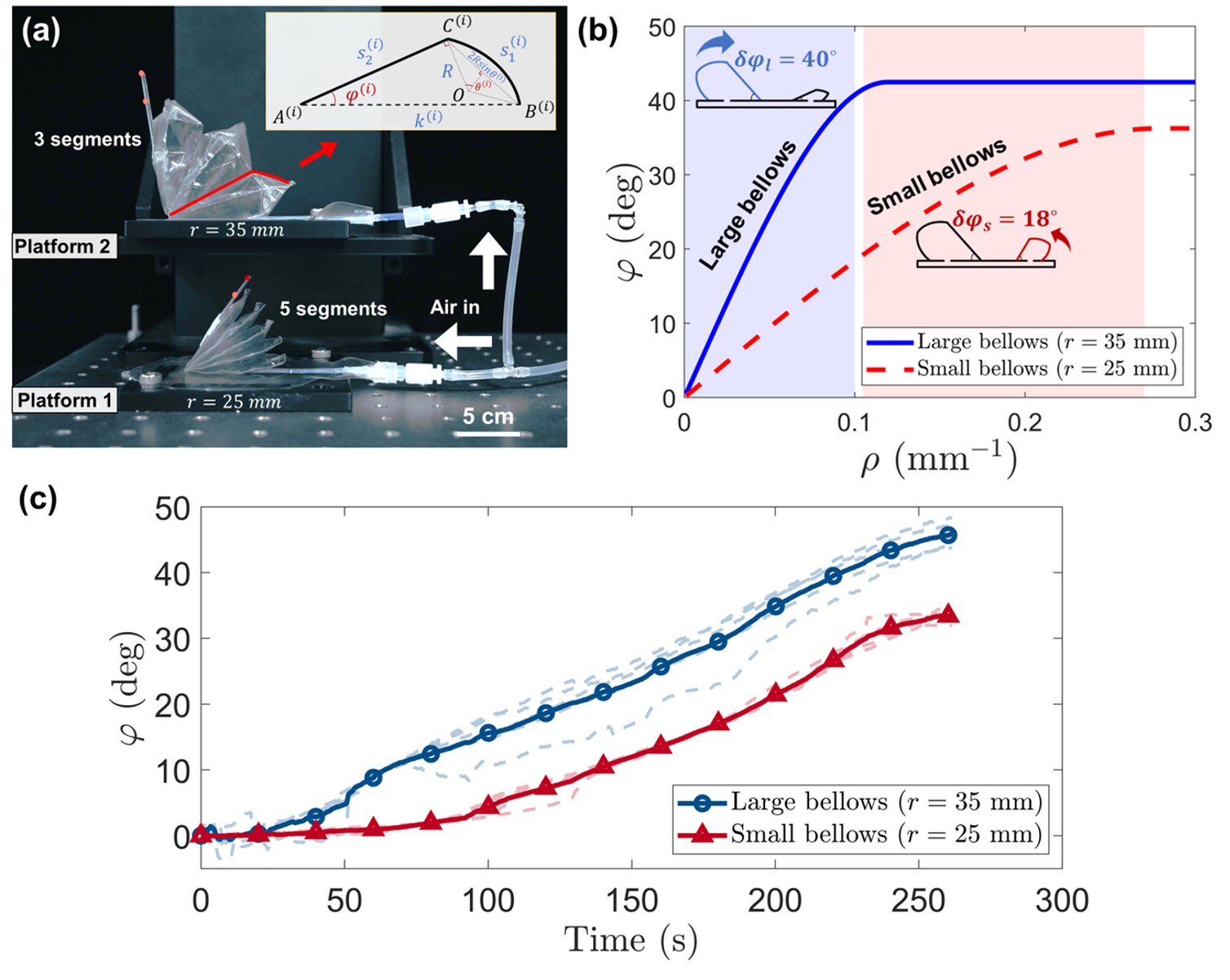}
  \caption{Experimental platform and modeling (a), model results (b), and experimental results (c) for the actuation sequencing of different radii in the output of the R-BESTS.} \label{Fig:6}
\end{figure}

\begin{figure*}[b]
  \centering
  \includegraphics[totalheight=4in]{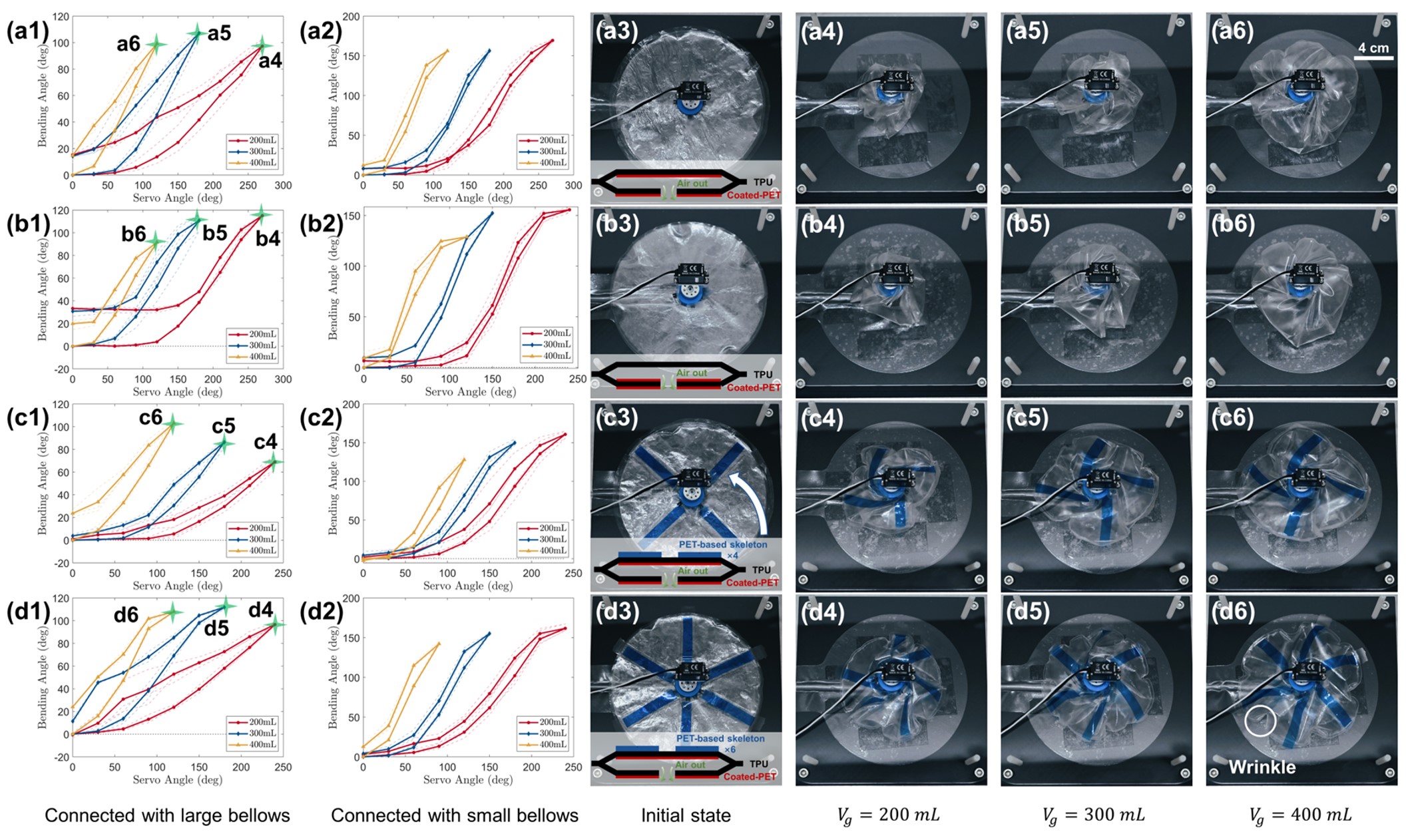}
  \caption{(a1)-(d1) and (a2)-(d2) indicate that the input bellows unit is connected to the large and small bellows units, respectively. (a3)-(d3) show the structure diagram of each input bellows unit. The states (a4)-(d4), (a5)-(d5), and (a6)-(d6) correspond to conditions where the large bellows unit is twisted to its limit, with initial gas volumes of \SI{200}{\milli\liter}, \SI{300}{\milli\liter}, and \SI{400}{\milli\liter}, respectively.} \label{Fig:7}
\end{figure*}
For each bending bellows unit, we observed the cross-sectional view and found that in the areas where the adjacent bellows segments were separated, the geometric shape was an arc due to the atmospheric pressure. As shown in Fig.\ref{Fig:6}(a), the central angle $2\theta^{(i)}$ corresponding to the arc is:
\begin{equation}
    2\theta^{(i)}=\frac{s_1^{(i)}}{R}=s_1^{(i)}\times \rho,
\end{equation}
where $R$ is the radius of curvature. The shape is approximately a straight line in the areas where the adjacent bellows segments are contacted. The length of the arc segment $s_1^{(i)}$ and the length of the straight line $s_2^{(i)}$ satisfy the following relationship:
\begin{equation}
    s_1^{(i)}+s_2^{(i)}=r^{(i)},
\end{equation}
where $r^{(i)}$ is the radius of TPU film when making the bending bellows unit as shown in Fig.\ref{Fig:3}(a).

Considering the status of bending bellows segments that connected with different $r^{(i)}$, we analyzed the relationship between their bending angles and the curvature. Since the internal pressure is uniform, bellows segments with different sizes exhibit the same curvature $\rho$. We established a geometric model as shown in Fig.\ref{Fig:6}(a).

\begin{equation}
\resizebox{\linewidth}{!}{$
    k^{(i)}=\sqrt{{s_2^{(i)}}^2+\left(2 R \sin \theta^{(i)}\right)^2-2 \cdot s_2 \cdot\left(2 R \sin \theta^{(i)}\right) \cdot \cos \left(\pi-\theta^{(i)}\right)},
    $}
\end{equation}

\begin{equation}
    \varphi^{(i)}=\arcsin \frac{2 R \sin \theta^{(i)} \cdot \sin \left(\pi-\theta^{(i)}\right)}{k^{(i)}},
\end{equation}
where $k^{(i)}$ is the length of $A^{(i)}B^{(i)}$ and $\varphi^{(i)}$ is the bending angle of half bellows segment.

The model results are shown in Fig.\ref{Fig:6}(b). As the curvature increases from zero, the amount of gas inside the R-BESTS gradually increases, and both the large and small bellows segments begin to bend. The large bellows segment has a higher bending rate and thus reaches its maximum bending angle faster than the smaller ones. 
To verify the model, we conducted the experiments (Fig.\ref{Fig:6}(a)) in which two TPU bending bellows units with different radii were inflated at a constant flow rate.

A digital camera was positioned laterally to provide a clear view of the bellows. It was placed on a stable tripod at a fixed distance to ensure consistent framing across all experiments.
Small adhesive markers were placed at the top of each bellows unit.
We used an open-source image analyzing
software (Tracker, http://physlets.org/tracker/) to measure the bending angle of the bellows unit.
Each bellows unit was securely attached to a rigid base plate to ensure consistent reference points.
Detailed experimental results are shown in Fig.\ref{Fig:6}(c).
The dashed lines represent the data from five experiments, and the solid lines represent the average of the five experiments. 

The experimental results confirm that the large bellows segments tend to bend earlier than the small ones. This difference arises because the large bellows segments are more flexible, requiring less pressure to initiate bending. In contrast, small bellows segments need a higher initial pressure to overcome internal resistance before bending begins due to their relatively higher rigidity.
Based on this property, we can achieve the actuation sequencing of the bending bellows units through structural design, thereby controlling the different gaits of the robot.

\subsection{Twisting of the input bellows unit}
To reveal the relationship between the input servo angle and the output bending angle, we designed four different structures for the input bellows units (Fig.\ref{Fig:7}(a3-d3)). 
The primary goals of the experiments were: (1) verifying the servo's ability to drive fluid volume changes within the bellows unit; (2) recording the bending of the bellows resulting from the servo's rotation angle; (3) quantifying the system's repeatability and reliability over multiple actuation cycles.
We verified baseline servo angles and initial fluid volumes (\SI{200}{\milli\liter}, \SI{300}{\milli\liter}, and \SI{400}{\milli\liter}) before each test to maintain consistent starting conditions. The servo was driven through a predefined step input, causing the cyclic bending angle of the bellows. Essential metrics included the bending angles of the large and small bellows. We repeated each test three times under identical conditions to evaluate the statistical variation. We also used the open-source software Tracker for image-based measurements here. 

We used an identical laminate (TPU and coated-PET) for the top and bottom films of the input bellows unit in Fig.\ref{Fig:7}(a).
Fig.\ref{Fig:7}(a1-a2) shows that the bending angle of the output is smaller when the servo is rotated to the same angle compared with other structures. 
Fig.\ref{Fig:7}(b) shows that the top and bottom films of the input bellows unit are different: a single TPU film for the top layer and a PET-TPU-PET laminate for the bottom layer. Fig.\ref{Fig:7}(b3-b6) shows the twisted geometry becomes random and disordered. 

\begin{figure*}[htbp]
  \centering
  \includegraphics[totalheight=4.2in]{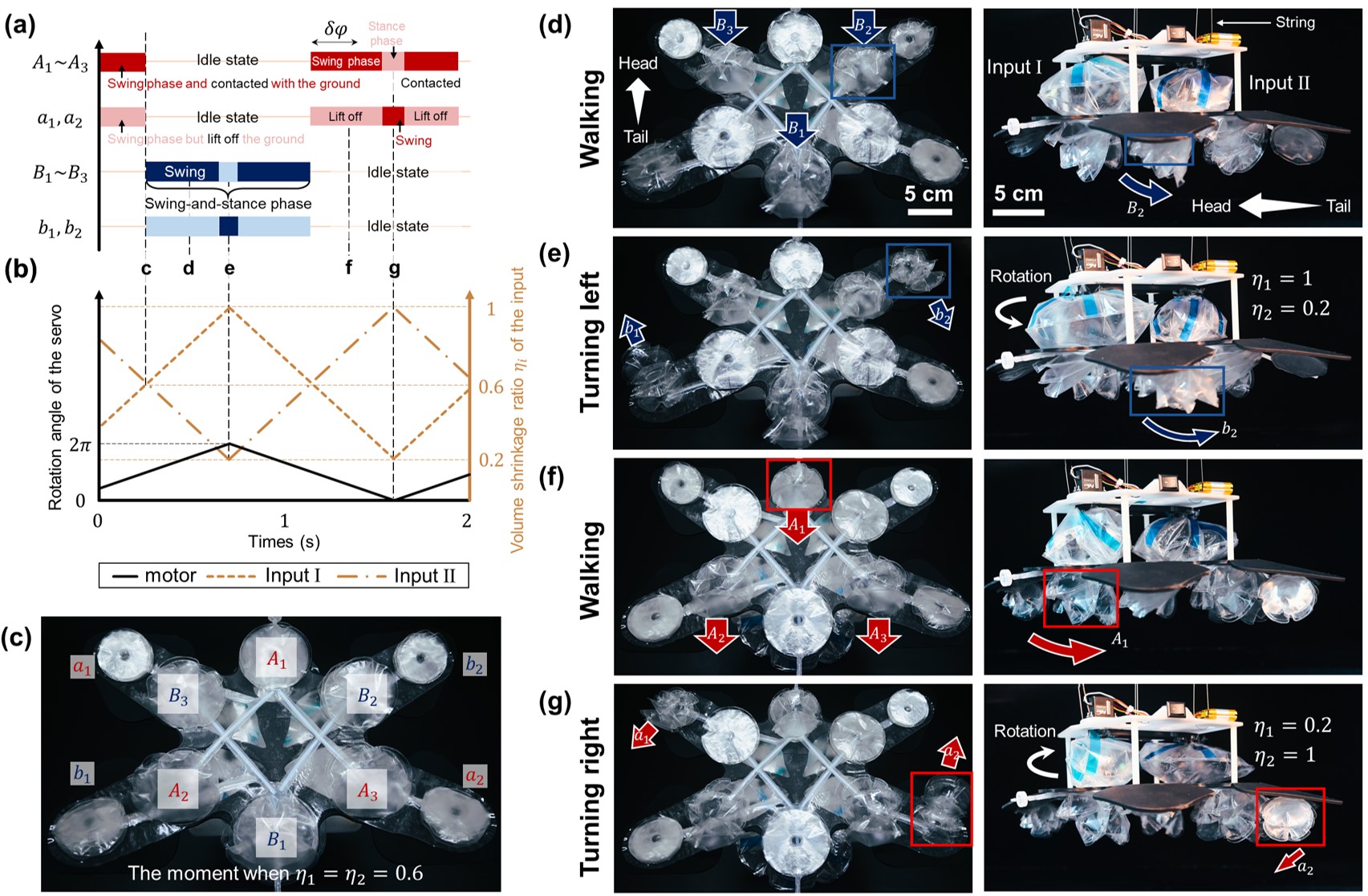}
  \caption{(a) Locomotion strategy and control for the alternating-tripod gait. Each bar in the diagram represents the swing phase of each leg. (b) Phase diagram of the servo and the two input bellows units. (c) Bottom view of the robot's ten legs. Schematic diagram of the robot walking (d), turning left (e), walking again (f), and turning right (g) in one cycle.} \label{Fig:8}
\end{figure*}

Therefore, we introduced a PET-based skeleton in Fig.\ref{Fig:7}(c) and (d), and this modification enhanced the lifetime of the input bellows unit. We used identical laminates for the top and bottom layers (the same as Fig.\ref{Fig:7}(a) of the input bellows unit). The bending speed of the six-rib skeleton (Fig.\ref{Fig:7}(d)) at the output is greater than that of the four-rib skeleton (Fig.\ref{Fig:7}(c)).
A larger number of skeletons leads to a greater squeezing area at the same servo twist angle. 
In addition, Fig.\ref{Fig:7}(d) shows that a pleat appeared between two adjacent skeletons, promoting the orderly shrinkage of the input bellows unit.

\subsection{Locomotion strategy and control}

Fig.\ref{Fig:8} shows how the robot achieves different gaits of walking and turning driven by only one servo. As shown in Fig.\ref{Fig:5}(b) and Fig.\ref{Fig:8}(c), each input bellows unit is connected to five bending bellows units, including three large bellows units ($A_1\sim A_3$, $B_1\sim B_3$) and two small bellows units ($a_1, a_2, b_1, b_2$).
The two input bellows units are in opposite phases due to the timing belt (Fig.\ref{Fig:8}(b)).
The ratio of the swing-and-stance phase is $50\%$ in one cycle. In each gait cycle, the legs connected with the same input bellows unit bend in the same phase.

As shown in Fig.\ref{Fig:8}(c), the states of group A and B exchange when the volume shrinkage rates $\eta$ of the two input bellows units are the same. In the next gait cycle, group B first entered the swing-and-stance phase, the large bellows units ($B_1\sim B_3$) first contacted the ground while the small bellows units ($b_1, b_2$) lifted off the ground. The robot was capable of achieving the walking gait (Fig.\ref{Fig:8}(d)). Then, the small bellows units ($b_1, b_2$) contacted the ground, leading the robot to turn left (Fig.\ref{Fig:8}(e)). At this stage, the large bellows units ($B_1\sim B_3$) were in the stance phase. When the servo rotated in the reverse direction, the bellows units in group A transited from the idle state to the swing-and-stance phase. Similarly, walking (Fig.\ref{Fig:8}(f)) and turning (Fig.\ref{Fig:8}(g)) motions were performed sequentially.


Notably, we establish a mapping relationship between the servo rotation angle and the robot's motion state. During the robot's movement, we do not constrain its motion to follow the period shown in Fig.\ref{Fig:8}, where the servo phase ranges from $0$ to $2\pi$. By controlling the phase position of the motor, we can not only control the robot to continuously accomplish left ($\frac{2\pi}{3}\sim\pi$) and right turning ($\pi\sim\frac{4\pi}{3}$) and walking ($0\sim\frac{2\pi}{3}$ or $\frac{4\pi}{3}\sim2\pi$) but also design the trajectory of the robot.

\section{EXPERIMENTS AND RESULTS}

\subsection{Basic Motion Performance}
The basic motions of the decapod robot were tested and verified based on the control strategy (Fig.\ref{Fig:8}). All experimental results are included in the accompanying videos.

\begin{figure}[htbp]
  \centering
  \includegraphics[totalheight=2.5in]{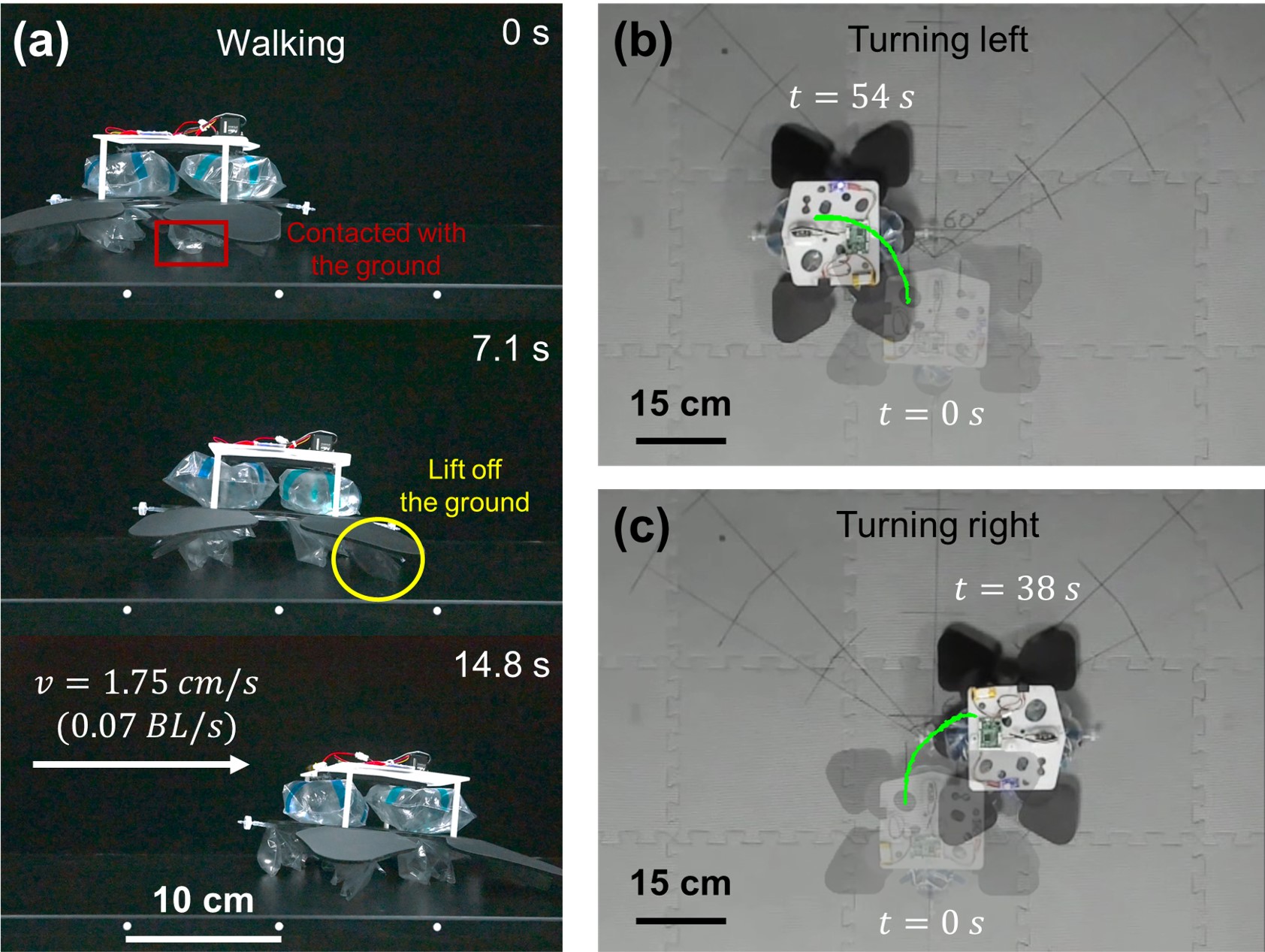}
  \caption{Basic motion performance of the robot: walking (a), turning left (b) and right (c).} \label{Fig:9}
\end{figure}

\subsubsection{Walking}
Fig.\ref{Fig:9}(a) shows a video snapshot of the robot during the experiment (8 cycles) on a flat marble tabletop. The experimental settings were as follows: each R-BESTS had a gas volume of \SI{600}{\milli\liter}, the phase position of input servo was $0\sim\frac{2\pi}{3}$, and the period of one cycle was \SI{2.8}{\second}. As shown in Fig.\ref{Fig:9}(a), the robot traveled a body length in \SI{14.8}{\second}, with an average speed of 0.07 BL/s.

It is worth noting that this is relatively fast for an untethered, soft-legged robot driven by one single input compared with other crawling robots under the same actuated condition\cite{r26,r12}.

\subsubsection{Turning}

Fig.\ref{Fig:9}(b) shows a top view of the robot turning on a foam cushion. The gas volume of each R-BESTS was consistent with the walking experiment. However, the phase positions of servo were $\frac{2\pi}{3}\sim\pi$ and $\pi\sim\frac{4\pi}{3}$ for the left and right turning, respectively.

Specifically, the period of the servo in the left-turn experiment was \SI{4}{\second}, and the period of the right-turn was \SI{3}{\second}. The corresponding times for the robot to rotate 90 degrees were \SI{54}{\second} and \SI{38}{\second}, respectively.

\begin{figure}[htbp]
  \centering
  \includegraphics[totalheight=3in]{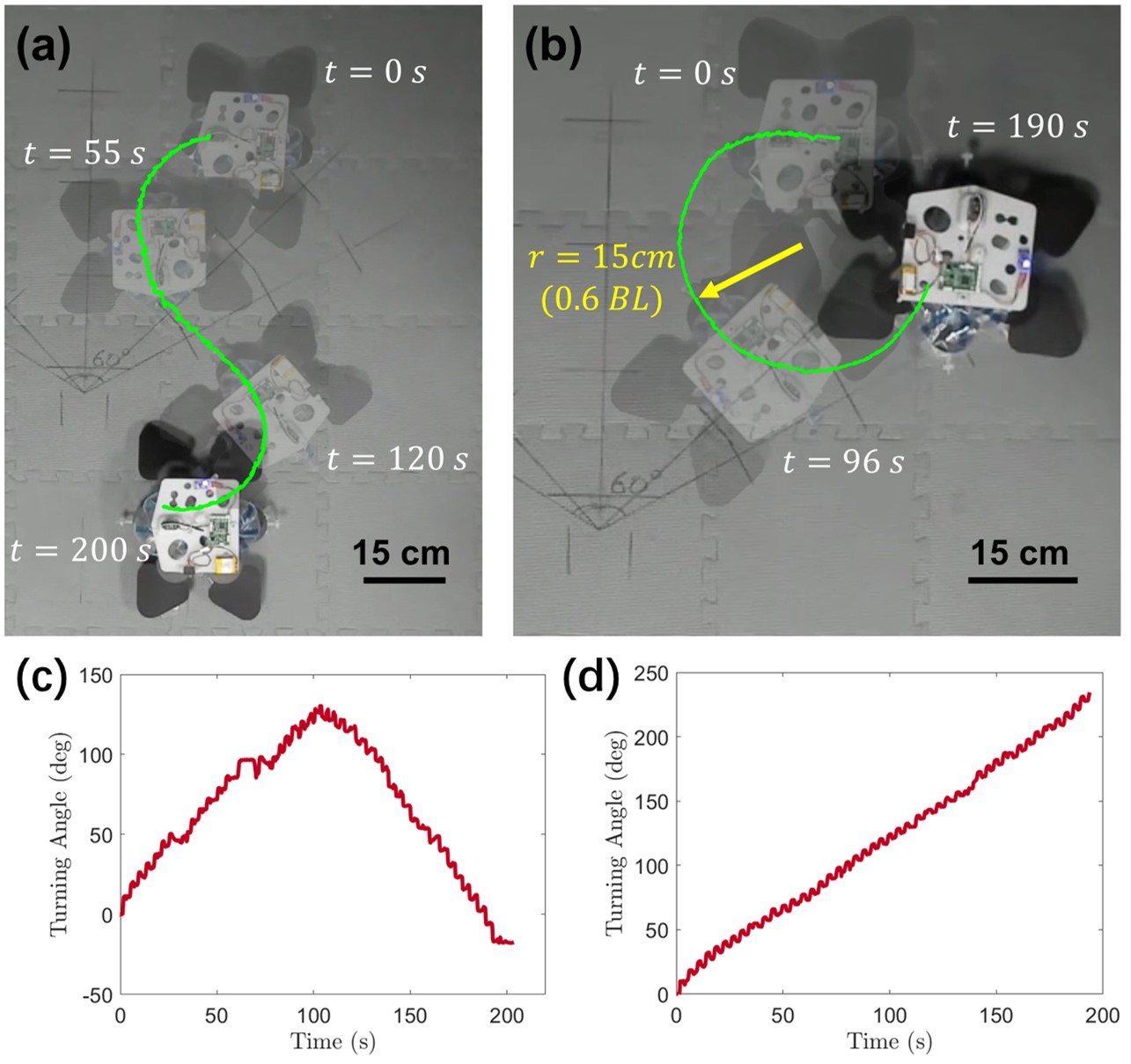}
  \caption{The robot following an S-shaped (a) and O-shaped curve (b). The turning angle of the robot during the S-shaped path (c) and O-shaped path (d).} \label{Fig:10}
\end{figure}

\subsection{Demonstration of locomotion skills}
\subsubsection{Path following}
By combining the previous three gaits (walking, turning left, and turning right), the robot could move along a predetermined trajectory. As shown in Fig.\ref{Fig:10}(a) and (b), the robot completed the S-shaped curve in \SI{200}{\second} and the O-shaped curve in \SI{190}{\second}, respectively. Fig.\ref{Fig:10}(c) and (d) show the angle between the orientation of the robot during movement and the initial moment.

\subsubsection{Crawling on different terrains}

We selected four representative types of complex terrain to verify the environmental robustness of the robot, including a cork board, a soft silicone foam mat, climbing a slope, and turning when carrying a payload.

\begin{figure}[htbp]
  \centering
  \includegraphics[totalheight=3in]{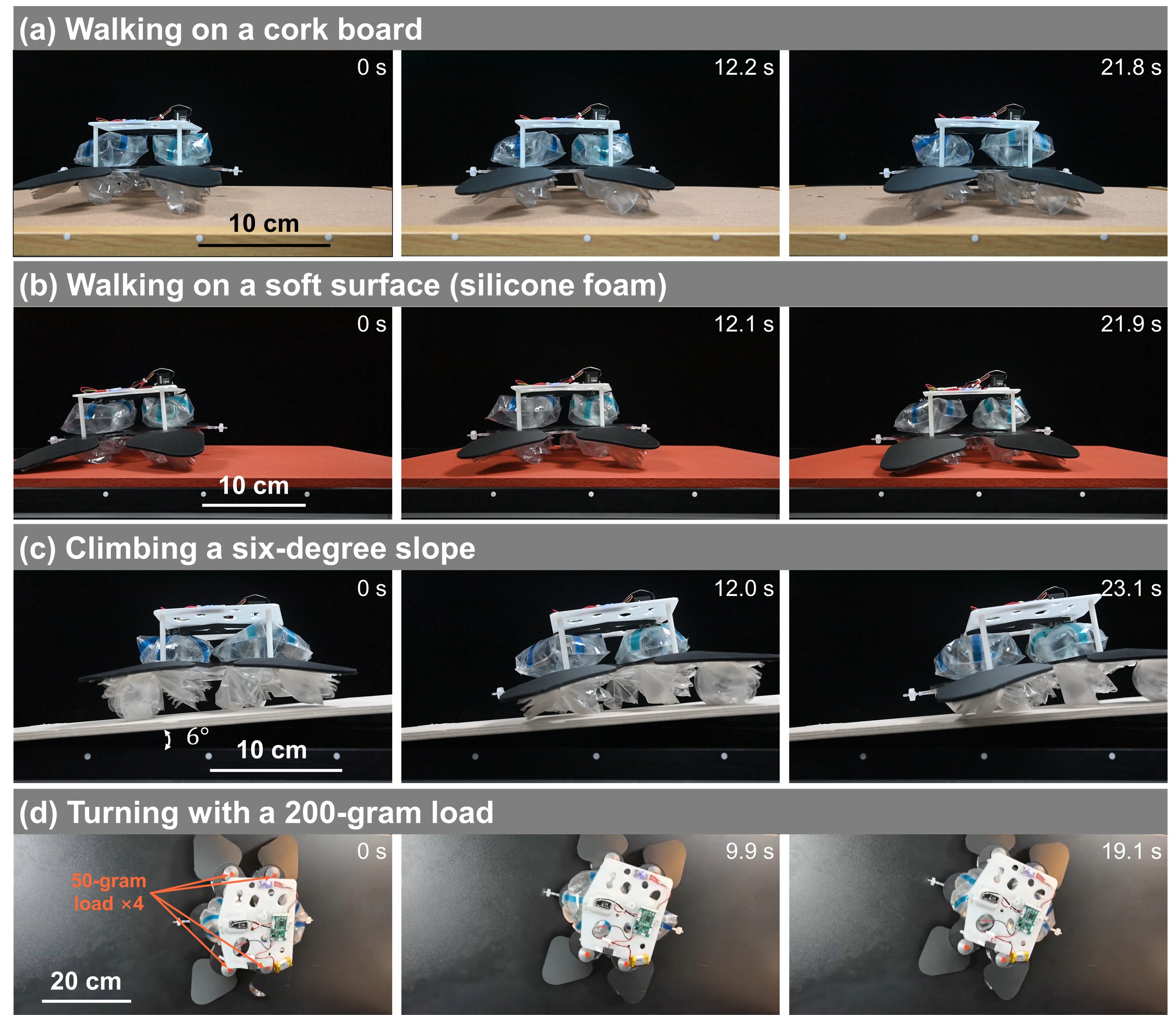}
  \caption{Snapshots of the robot when crawling on different terrains. (a) A corkboard. (b) A soft surface with silicone foam. (c) Climbing a slope. (d) Turning with a payload.} \label{Fig:11}
\end{figure}

Fig.\ref{Fig:11}(a) and (b) show the experiment of the robot walking in a straight line on a cork surface and a soft silicone foam. In Fig.\ref{Fig:11}(a), due to the rough surface of the wood, the robot's walking speed was 0.023 BL/s. We found that some legs of the robot couldn't rebound after bending on the soft surface shown in Fig.\ref{Fig:11}(b), and the deformation of the soft ground led to increased resistance, resulting in a speed of 0.019 BL/s. 
Although the speeds in these two terrains were lower than those on a flat surface, it demonstrated the soft robot's ability to move despite the failure of some legs.

The ability of slope climbing was evaluated through an oblique plane using a flat rigid plate (Fig.\ref{Fig:11}(c)). The slope angle was set to $6^\circ$, and experimental results show that the robot's speed was 0.018 BL/s. This climbing ability endows a higher level of robustness. 

In the payload carrying capacity test shown in Fig.\ref{Fig:11}(d), the robot could carry a \SI{200}{\gram} load ($59.9\%$ of its own weight) at the speed of 0.020 BL/s on flat rigid terrain with the same turning radius compared with zero payload.

\section{DISCUSSION AND CONCLUSION}

In this study, we have proposed a soft-legged decapod robot driven by a single servo. A closed soft transmission system (R-BESTS) is implemented in this robot. A timing belt is used to constrain the phases of two input bellows units of the R-BESTS. Besides, we control the robot legs' bending sequence by predesigning the radius of output bellows units. The decapod robot is capable of walking (crawling) at 1.75 centimeters per second and turning with a 15-centimeter radius. The properties and performances of the robot are shown in Table \ref{Table}. 

\begin{table}[htbp]
\centering
\caption{Properties and performance of our robot}\label{Table}
\begin{tabular}{ll}
\hline
Parameters             & Values \\ \hline
Weight          &   334 g   \\
Dimensions     &  25 cm$\times$36 cm$\times$10 cm   \\
Load capacity  &  200 g     \\
Speed          &  1.75 cm/s (0.07 BL/s)     \\
Turning radius  &  15 cm (0.6 BL)    \\
Operation time  &   90 min    \\ \hline
\end{tabular}
\end{table} 

A significant benefit of our R-BESTS is that the servo motor operates without direct contact with the internal fluid. This configuration helps reduce friction-related power losses. In addition, eliminating valves in the system reduces mechanical complexity and decreases the risk of gas leakage.
Together, these factors can lead to lower overall energy consumption, thereby improving the efficiency of the servo-driven system.
In addition, the R-BESTS exhibits lower noise levels and reduces vibration because it does not rely on continuous compression of gas.

However, the current bellows structures of the R-BESTS are designed empirically, and we have not thoroughly analyzed the occurrence rules and specific functions of those structural wrinkles on the twisted bellows. To achieve better transmission performances and more elegant torsional topologies, we will introduce computational design methods to further optimize the design parameters of the input bellows units and their skeletons, such as the position, shape, and thickness. 

\bibliographystyle{ieeetr}
\bibliography{ref.bib}

\end{document}